\documentclass{acmart}

\AtBeginDocument{%
  }

\copyrightyear{2024}
\acmYear{2024}
\setcopyright{rightsretained}
\acmConference[AM '24]{Audio Mostly 2024 - Explorations in Sonic Cultures}{September 18--20, 2024}{Milan, Italy}
\acmBooktitle{Audio Mostly 2024 - Explorations in Sonic Cultures (AM '24), September 18--20, 2024, Milan, Italy}
\acmPrice{}
\acmDOI{10.1145/3678299.3678324}
\acmISBN{979-8-4007-0968-5/24/09}
\usepackage{graphicx}

\begin{document}

\title{Optical Music Recognition in Manuscripts from the Ricordi Archive}

\author{Federico Simonetta}
\orcid{0000-0002-5928-9836}
\affiliation{%
  \institution{GSSI - Gran Sasso Science Institute}
  \city{L'Aquila}
  \country{Italy}
}
\email{federico.simonetta@gssi.it}

\author{Rishav Mondal}
\orcid{0009-0008-4101-0906}
\email{rishav.mondal @ studenti.unimi.it}
\author{Luca Andrea Ludovico}
\orcid{0000-0002-8251-2231}
\email{luca.ludovico @ unimi.it}
\author{Stavros Ntalampiras}
\orcid{0000-0003-3482-9215}
\email{stavros.ntalampiras @ unimi.it}
\affiliation{%
  \institution{University of Milan}
  \city{Milan}
  \country{Italy}}

\renewcommand{\shortauthors}{Simonetta et al.}

\begin{abstract}
The Ricordi archive, a prestigious collection of significant musical manuscripts from renowned opera composers such as Donizetti, Verdi and Puccini, has been digitized. This process has allowed us to automatically extract samples that represent various musical elements depicted on the manuscripts, including notes, staves, clefs, erasures, and composer's annotations, among others. To distinguish between digitization noise and actual music elements, a subset of these images was meticulously grouped and labeled by multiple individuals into several classes. After assessing the consistency of the annotations, we trained multiple neural network-based classifiers to differentiate between the identified music elements. The primary objective of this study was to evaluate the reliability of these classifiers, with the ultimate goal of using them for the automatic categorization of the remaining unannotated data set.
The dataset, complemented by  manual annotations,  models, and source code used in these experiments are publicly accessible for replication purposes.\footnote{Code repository: \url{https://github.com/LIMUNIMI/RicordiArchiveOMR}; dataset repository: \url{https://zenodo.org/doi/10.5281/zenodo.11186095}}
\end{abstract}



\begin{teaserfigure}
  \centering
  \includegraphics[trim={0 30cm 0 0},clip,width=\textwidth]{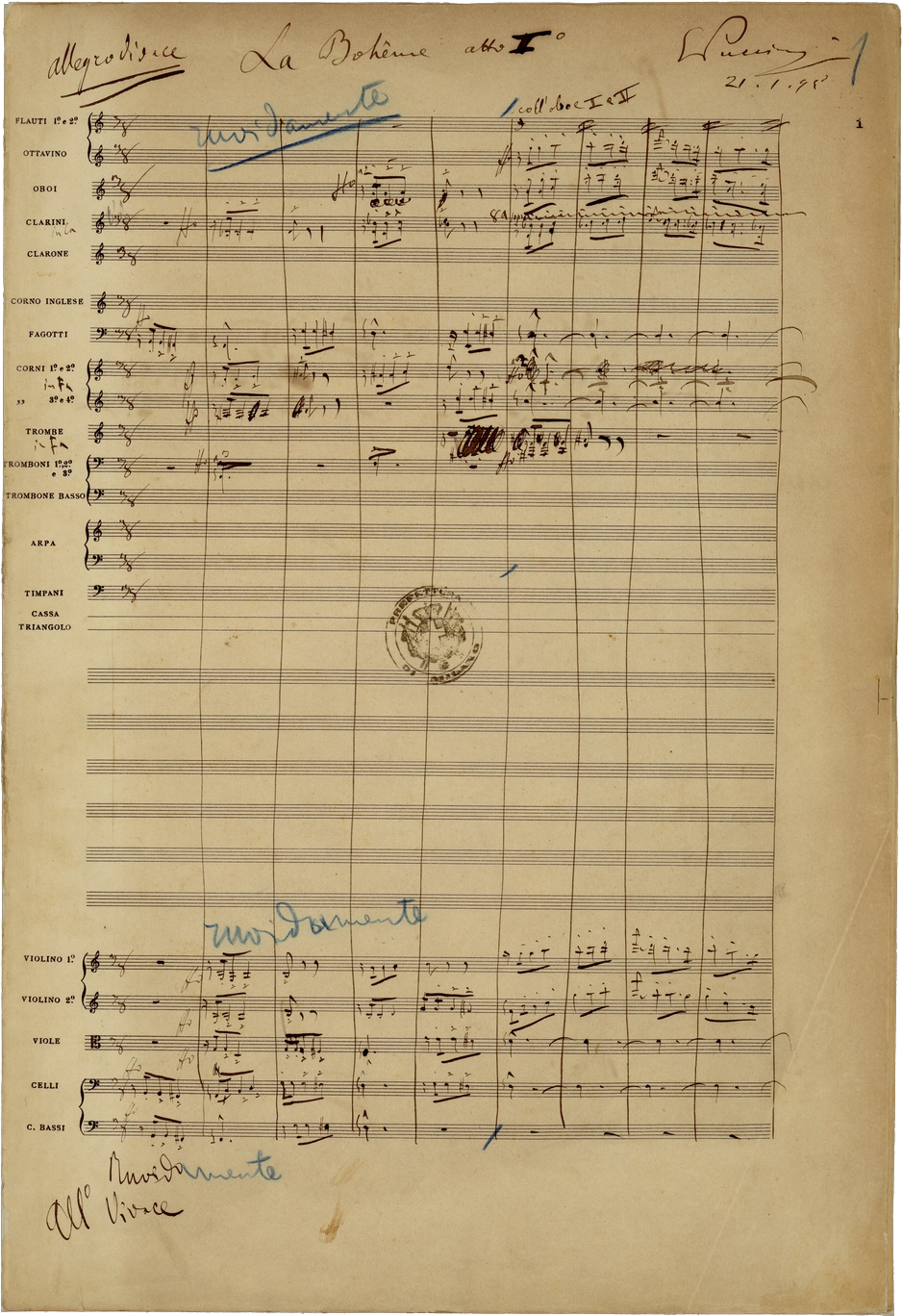}
  \caption{Example of handwritten score from the Archivio Storico Ricordi from the manuscript score of ``La Bohème'', by Giacomo Puccini.}
  \Description{Example of handwritten score from the Archivio Storico Ricordi from the manuscript score of ``La Bohème'', by Giacomo Puccini.}
  \label{fig:archive}
\end{teaserfigure}


\maketitle

\section{Introduction}

The Ricordi Archive, or \textit{Archivio Storico Ricordi} in Italian, is a vast repository of historical documents amassed by the Italian publisher, Ricordi. The archive is renowned for its digitized manuscripts of distinguished opera composers such as Donizetti, Verdi, and Puccini. These manuscripts, digitized and systematically cataloged in a database maintained by the author's institution,\footnote{https://ricordi.lim.di.unimi.it} are a significant asset for musicological and historical research.

This study aims to annotate the entire database with pertinent musical symbols, thereby improving the accessibility and discoverability of these priceless manuscripts. To achieve this, we developed a suitable Optical Music Recognition (OMR) methodology. 

OMR is a subfield of computer science dedicated to converting music notation from a visual format, such as scanned images or printed music sheets, into a digital form that can be manipulated by software. It serves as a bridge between physical music representations and their digital equivalents, with the aim of automating the interpretation of music symbols for various applications, including content retrieval, digital music libraries, and musicological research~\cite{calvo-zaragozaUnderstandingOpticalMusic2020,9733531,SimoISMIR2019}.

The evolution of OMR, particularly for printed music, is largely attributed to advancements in image processing and machine learning. Although early efforts relied heavily on rule-based systems~\cite{fujinagaAdaptiveOpticalMusic1997}, contemporary strategies use modern neural architectures. These architectures excel in feature extraction and help to accurately identify and classify musical symbols across various datasets and notation styles~\cite{CalvoZaragoza2017RecognitionOH,GarridoMunoz2022AHA,liTrOMRTransformerBasedPolyphonic2023}.

Handwritten Music Recognition (HMR) focuses on the recognition of handwritten scores, adding an extra layer of complexity due to the unique styles and nuances inherent in individual handwriting. Recent methodologies suggest various strategies, including data augmentation and transfer learning, to enhance system performance and address the challenges specific to HMR~\cite{calvo-zaragozaUnderstandingOpticalMusic2020,Zhang2023ADF}.

Automating the analysis of handwritten music presents significant obstacles due to the intricacy and variability of human handwriting; here, progress has been driven by the implementation of advanced machine learning models, specifically Convolutional Recurrent Neural Networks. Such models effectively capture both the spatial characteristics of the image and the sequential nature of music notation, which are crucial for the development of successful OMR systems for handwritten scores~\cite{baroOpticalMusicRecognition2019}.

The development of large-scale datasets has been fundamental to the advancement of OMR technology. A key resource in OMR research is the MUSCIMA++ dataset~\cite{Hajic2017TheMD}. This dataset, comprising 140 pages of handwritten music scores, is meticulously annotated with over 91000 symbols across 107 classes. It facilitates a wide range of tasks, including symbol classification and notation graph assembly. MUSCIMA++ is an extension of the CVC-MUSCIMA dataset, which includes 1,000 music sheets from 50 musicians. The detailed annotations of musical symbols and their interrelations in MUSCIMA++ are crucial for training and evaluating OMR systems. Its extensive coverage of musical notations, from notes to articulation marks, and structural annotations detailing symbol relationships, make it an invaluable tool for machine learning applications in music symbol detection and recognition. Moreover, MUSCIMA++ serves as a benchmark in the OMR field, contributing significantly to the development of technologies that convert sheet music into digital formats, thereby enhancing the accuracy and reliability of OMR systems.

The DeepScores~\cite{Tuggener2020TheDD} dataset is another valuable resource, consisting of high-quality images of printed music divided into approximately 300,000 sheets of musical scores with nearly a hundred million small objects. It provides ground truth for object classification, detection, and semantic segmentation, focusing on the recognition of small objects and intricate details in musical scores.

Carefully curated and annotated datasets, such as MUSCIMA++, are instrumental in advancing the field of HMR. They not only improve the accuracy of recognition models, such as Convolutional Recurrent Neural Networks, but also establish rigorous benchmarks for evaluating these models against the complexities of real-world musical notations. By offering a comprehensive collection of samples that cover the full range of human handwriting variability, these datasets are essential to refine the capabilities of end-to-end OMR systems. Consequently, these systems can accurately interpret the intricate nuances of handwritten music scores, addressing a significant academic and practical challenge.

In this study, we address a series of the challenges in OMR by providing a new dataset of musical symbols from real-world annotated manuscripts, and by training and evaluating several neural classifiers to distinguish between these symbols. Moreover, we expect that the current work will be an important step towards the automatic annotation of the entire Ricordi Archive. The data set is published online.\footnote{\label{link:zenodo}\url{https://zenodo.org/doi/10.5281/zenodo.11186095}}

The rest of the paper is structured as follows: in Sec.~\ref{sec:background} we provide an insight into the history and the activities of the Ricordi Archive. Sec.~\ref{sec:dataset} presents the dataset of images focusing on preprocessing and annotation processes, while in Sec.~\ref{sec:experiments} we conduct a series of experiments assessing the ability of statistical models in music symbol identification. Sec.~\ref{sec:results} discusses the obtained results, and, finally, in Sec.~\ref{sec:conclusions} we draw the conclusions.

\section{Background}
\label{sec:background}
% \begin{figure}
%   \centering
%   \includegraphics[trim={0 30cm 0 0},clip,width=\textwidth]{imgs/boheme}
%   \caption{Example of handwritten score from the Archivio Storico Ricordi from the manuscript score of ``La Bohème'', by Giacomo Puccini.}
%   \Description{Example of handwritten score from the Archivio Storico Ricordi from the manuscript score of ``La Bohème'', by Giacomo Puccini.}
%   \label{fig:archive}
% \end{figure}

The Ricordi Archive originated alongside the publishing house Casa Ricordi, established in 1808. Regarded as a paramount private musical repository, it safeguards the original handwritten scores of 23 out of Verdi's 28 operas, all operas by Giacomo Puccini (except \textit{La Rondine}), and numerous works by composers such as Bellini, Rossini, Donizetti, as well as contemporary composers like Nono, Donatoni, Sciarrino, and Bussotti.

The archive's exceptional significance lies in the diversity of its materials, offering an articulated view of Italian culture, industry, and society. This archive preserves an extensive collection of visual materials associated with numerous premieres worldwide and locally, encompassing set and costume designs, photography compilations, correspondence, and business records. These resources empower researchers to reconstruct the inception of significant operas and the evolution of the musical publishing industry during the 19\textsuperscript{th} and early 20\textsuperscript{th} Centuries. 
Furthermore, the visual collection covers various artistic domains such as painting, stage design, and decorative arts, offering insights into costume history, jewelry design, stage properties, and the broader publishing landscape. It also sheds light on the relationship between publishers and artists across different fields and provides glimpses into the theatrical realm. Scholars can trace the personal and professional trajectories of numerous composers from their earliest works, such as Verdi's \textit{Oberto Conte di San Bonifacio} and Puccini's \textit{Le Villi}, to their most important operas like Verdi's \textit{Falstaff} and Puccini's unfinished \textit{Turandot}.

The Ricordi Archive preserves approximately 8,000 scores, over 16,000 letters exchanged among musicians, librettists, singers, and other stakeholders, around 10,000 set and costume designs, more than 9,000 librettos, 6,000 historical photographs, and a substantial collection of Art Nouveau and Art Deco posters crafted by prominent artists of the era.

The digitization initiative of the historical archive stemmed from a collaborative effort involving the Italian Ministry of Culture, the National Department of Archives and Libraries, the Italian Supervisory Council for Libraries and Cultural Institutions, Casa Ricordi, Biblioteca Nazionale Braidense, and the Laboratory of Music Informatics (\textit{Laboratorio di Informatica Musicale}, LIM) of the University of Milan. This project, initiated in 2006, adheres to the standards established by the Italian National Library Service (\textit{Servizio Bibliotecario Nazionale}, SBN), which is overseen by the Central Unified Catalogue Institute (\textit{Istituto Centrale per il Catalogo Unico}, ICCU). Given the archive's artistic and historical significance, its preservation is subject to the regulations and oversight of the Ministry of Culture.

\section{Dataset}
\label{sec:dataset}

The original core of the digitization campaign of Ricordi Archive consisted of about 3000 digitized images, mainly handwritten scores by Donizetti, Puccini, Verdi, and Respighi.

\subsection{Preprocessing}

The creation of the dataset necessitated preliminary processing to identify pertinent objects and reduce the annotation effort in its initial phase. This process entailed the following steps:

\begin{itemize}

  \item Staff Line Removal -- A neural autoencoder-based algorithm~\cite{Gallego2017138} was employed to identify staff lines. Given the distinct clarity of the staff lines in the 19th-century documents from the archive, this method proved highly effective. The staff lines, being printed rather than handwritten, were easily distinguishable from the musical symbols, thereby enhancing the reliability of this step;

  \item Blob Detection -- The Difference of Gaussians (DoG) method, as implemented by the \texttt{scikit-image} Python module~\cite{Lowe2004}, was utilized to identify the musical symbols in the images. We used $\sigma \in [10, 50]$ and a threshold of 0.1. Although this step does not guarantee the detection of all relevant objects in the images, it was tuned to be particularly sensitive to the ink regions. As a result, a large number of false positives were included to minimize the occurrence of false negatives, i.e., relevant objects not included in the dataset;

  \item Rescale and Save -- The grayscale images of the detected blobs were stored after rescaling their intensity values to $[0, 255]$.

\end{itemize}

The construction of our inter-referencing database, which utilizes JSON files, began with the collection of blob images. In total, 473,238 blobs were extracted. These images were then systematically stored to facilitate easy access and reference. 

The initial step in this process involved generating a grayscale image for each image in the original Ricordi Archive by removing the staff lines. Each of these grayscale images was then associated with a JSON file, which contained a reference to the original image, the path to the grayscale image without staff lines, and a list of JSON files associated with the blobs detected in the image.

Subsequently, a set of blobs was detected for each image from which staff lines had been removed. Each detected blob was stored as a grayscale image, and a JSON file was created for each blob. This file contained a reference to the parent image (the image without staves) and the bounding box of the blob. This systematic approach ensured that each blob and its associated data could be easily traced back to its parent image.

\subsection{Annotation}

\begin{figure}
    \centering
    \includegraphics[width=0.8\textwidth]{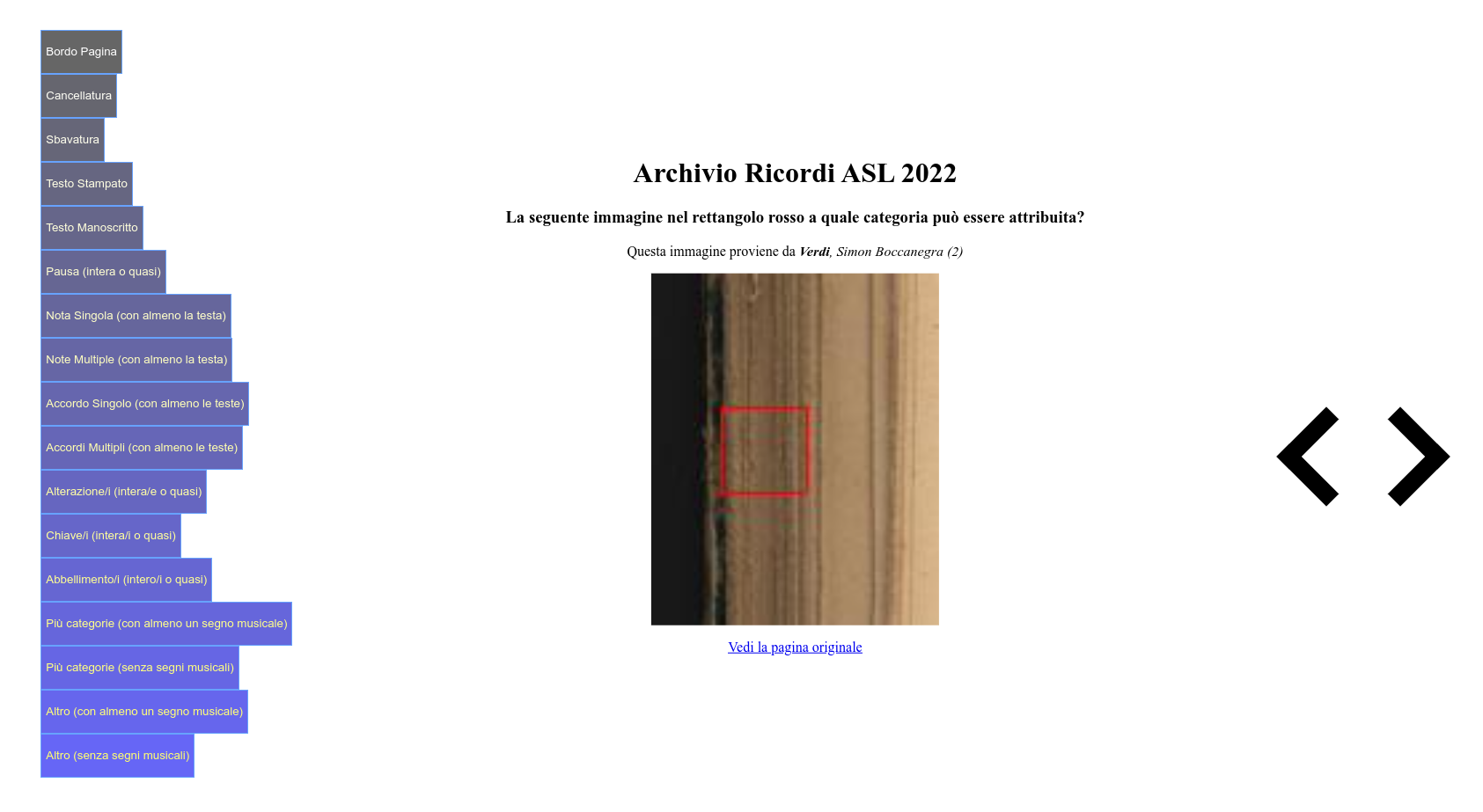}
    \caption{Screenshot of the interface used for annotating the dataset. Texts are in italian.}
    \Description{Screenshot of the interface used for annotating the dataset. Texts are in italian.}
    \label{fig:screenshot}
\end{figure}

The annotation phase was facilitated by 15 local high-school students with music reading skills, who were divided into seven groups of two or three.
We developed a custom interface that enabled the students to assign labels to each blob image.

For reference, each detected blob was highlighted with a bounding box within a larger excerpt of the original image. Additionally, an HTML link to the original image was provided for further examination if necessary. A screenshot of the annotation interface is depicted in Figure~\ref{fig:screenshot}.

We identified 16 classes of objects that could be recognized as blobs. These included page border, erasure, smudge, printed and handwritten text, rest, single or multiple notes, single or multiple chords, alterations, clefs, ornaments, multiple categories (with and without music signs), and an ``other'' category (with and without music signs) for objects that did not fit into any of the other categories.

To assess the accuracy of the annotations, a sample of 500 blobs was randomly selected for cyclic annotation by all annotators. This process involved the selection of a control blob with a 20\% probability at each annotation cycle. The control blobs were initially used during the annotation process to gamify the labeling work by providing the annotators with simple scores that reflect the quality of their work. The score was calculated based on the average of two factors: the Spearman correlation coefficient of the annotator's labels, which represents intra-agreement, and the Spearman correlation coefficient between a) the average of the annotator's labels for each control blob and b) the average of the annotations already stored in the database, provided by other annotators, representing inter-agreement.

The same control blobs were then used to compute the inter and intra-annotator agreement in order to assess the annotation quality. We first calculated a reference label for each control blob $i$ and annotator $j$ as the mode of the ratings given by annotator $j$ to the control blob $i$. On these sets of ratings, we computed Krippendorff's alpha (0.72), indicating that the raters generally agreed on the representation of the symbols. The intra-rater agreement was computed using Krippendorff's alpha over each annotator's labels separately, and was found to be between 0.52 and 0.71 depending on the annotator, indicating that the annotators were generally coherent on the representation of the symbols. 

\begin{figure}
    \centering
    \includegraphics[width=0.7\textwidth]{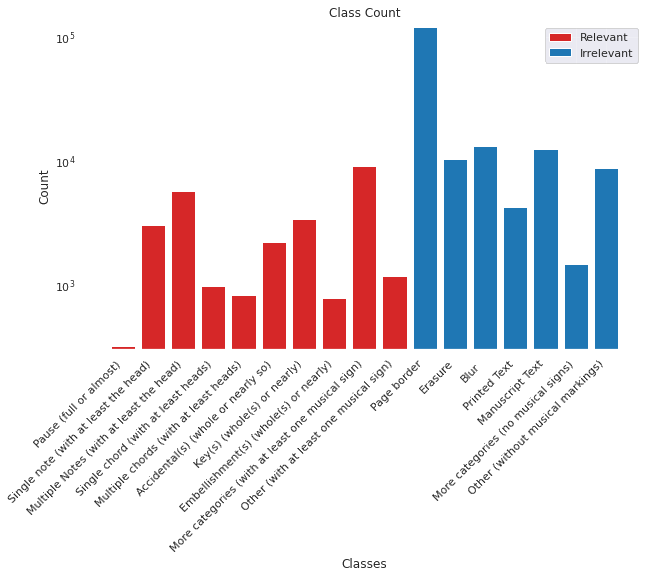}
    \Description{Original distribution of the images across the classes.}

    \caption{Original distribution of the blob images across the classes before merging the less frequent ones. Note that the Y axis is in log scale.}
    \label{fig:original-ditribution}
\end{figure}

To identify the reasons for the partial disagreement, we first identified a reference annotation for each control blob $i$ across all annotators using the mode of the labels given by all annotators. We then analyzed the normalized confusion matrix resulting from the annotated labels and the reference labels.
%We  each row of the confusion matrix to its maximum value. 
If a label was annotated correctly on average, the maximum value of each row was along the diagonal. The remaining values were then in reference to such value, so that a value near to 1 would mean that there was confusion among the annotators about the meaning of that label. We merged the classes that had a normalized confusion value greater than 0.5. We performed this step -- i.e. computation of the confusion matrix, normalization, and class merging -- iteratively until no classes were merged. This procedure led us to merge the ``multiple notes'' and ``multiple chords'' labels, as well as the labels ``blurs'' and ``multiple categories without musical signs'', thus resulting in 14 classes. The class merging increased Krippendorff's alpha to 0.84 for the inter-rater agreement, while the the intra-rater agreement raised to $[0.63, 0.79]$. The distribution is shown in Figure~\ref{fig:original-ditribution}.

\begin{figure}
    \centering
    \includegraphics[width=0.7\textwidth]{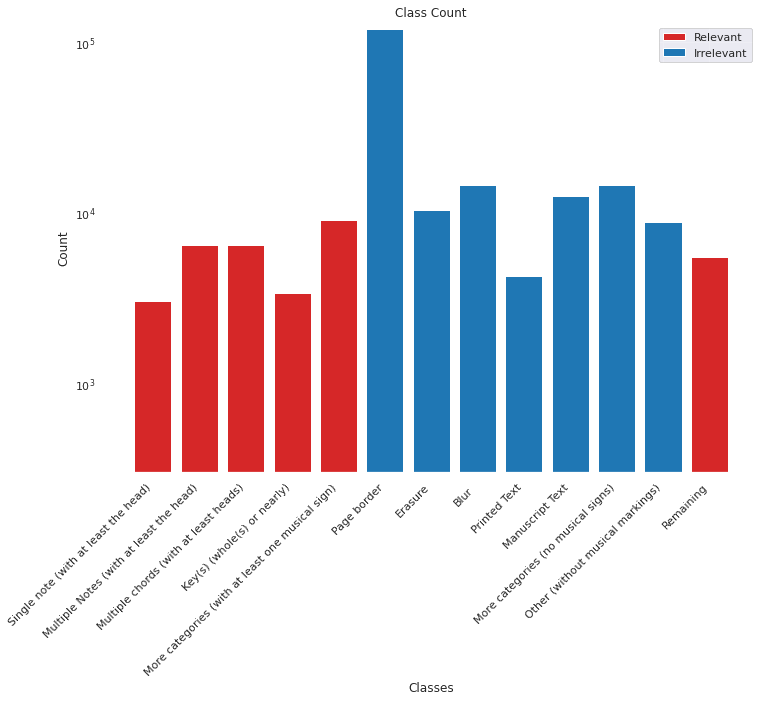}
    \Description{Distribution of the images across the classes, after the merging of the less frequent classes. Note that the Y axis is in log scale.}

    \caption{Distribution of the images across the classes, after the merging of the less frequent classes. Note that the Y axis is in log scale.}
    \label{fig:merged-distribution}
\end{figure}

To enhance the training of machine learning models, we addressed the issue of class imbalance by consolidating certain classes into a single category, termed as ``Remaining''. This amalgamation involved classes with sample sizes less than $0.75 \times m$, where $m$ represents the median class cardinality. Consequently, pauses, embellishments, single chords, accidentals, and ``Other (with musical signs)'' were merged, resulting in a total of 11 classes. The revised distribution of samples across these classes is depicted in Figure~\ref{fig:merged-distribution}.

For the purpose of simplifying the classification task, we categorized the labels into two clusters: ``musically relevant'' and ``musically irrelevant''. This distinction signifies the presence or absence of musical signs in the blob.
The inter-rater agreement for this binary annotation was measured using Krippendorff's alpha, yielding a value of 0.89. The intra-rater agreement ranged between 0.74 and 0.79.

\begin{figure}
    \centering
    \includegraphics[width=0.75\textwidth]{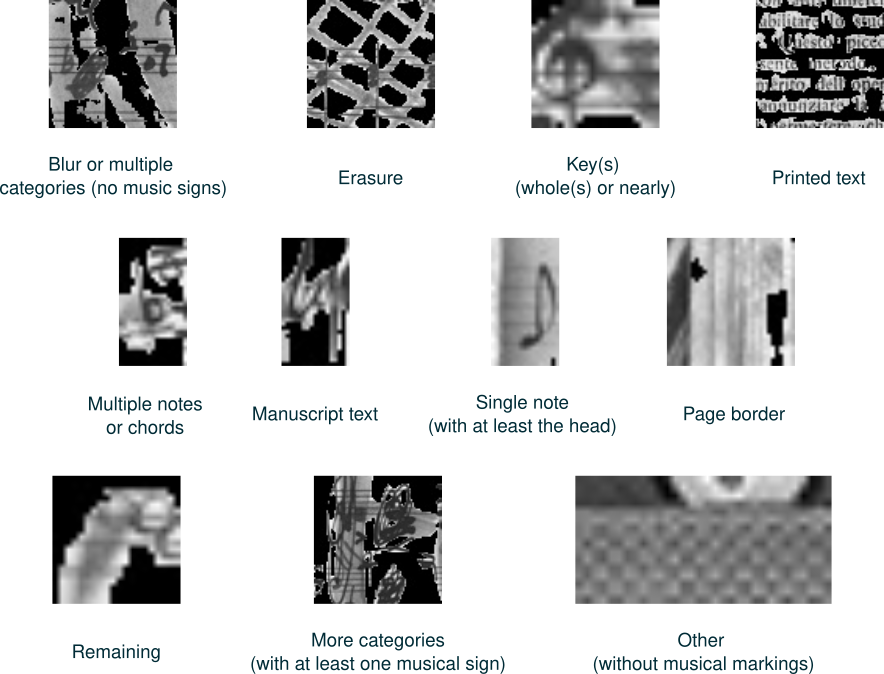}
    \caption{Examples of blobs for each class in the dataset.}
    \label{fig:blobs}
\end{figure}

The final class distribution in the proposed dataset, comprising a total of 198,159 annotated blobs, is illustrated in Figure~\ref{fig:merged-distribution}. Examples of blobs are shown in Figure~\ref{fig:blobs}. We further provide predefined splits for training and testing sets, applicable to both binary and multiclass classification tasks.

\section{Experiments}
\label{sec:experiments}

To evaluate the efficacy of statistical models in recognizing musical symbols, we fine-tuned three renowned deep learning classifiers: ResNet, DenseNet, and GoogleNet. We also employed an advanced AutoML method~\cite{feurerAutosklearnHandsfreeAutoML2022} to compare neural networks with conventional machine learning techniques.

Considering the significant imbalance depicted in Figure~\ref{fig:original-ditribution}, we initially subsampled the classes with the highest cardinalities to achieve perfectly balanced training and validation sets. This subsampling involved randomly selecting $n$ samples from the largest categories, where $n$ corresponds to the number of samples in the smallest category. While this method yields balanced training and validation sets, the test sets remain highly imbalanced. Therefore, it is crucial to implement appropriate validation measures that account for class imbalance during testing, as shown in  Tables~\ref{tab:binary}~and~\ref{tab:multiclass}.

In all instances, we enhanced the training set by applying random rotations of up to $\pm10$ degrees, random flips with probability of 0.5, brightness, contrast, and saturation jitters with factor 0.25. To improve the quality of the images, we also implemented Gaussian blur denoising with kernel of 3 and sigma 1.5 and contrast correction to 1.5 of the original contrast. To ensure compatibility with the ImageNet pre-trained weights, we renormalized the image channels and resized all images to $256 \times 256$ pixels.

For the deep-learning classifiers, we utilized the pre-trained weights available in the \texttt{torchvision} library, obtained from the ImageNet 1K dataset~\cite{torchvision2010,imagenet2009}. We re-trained all models using standard cross-entropy loss and the 1 cycle learning rate policy~\cite{smith2019super} with Stochastic Gradient Descent, setting the maximum learning rate at $0.01$. The models were trained for 500 epochs with early stopping based on validation loss, exhibiting a patience of 20 epochs. This resulted in approximately 40 epochs of actual training, with a maximum of 70 epochs for GoogleNet in binary classification. We used a batch size of 64 and allocated 68\%, 17\%, and 15\% of the dataset for training, validation, and testing, respectively.

We placed particular emphasis on uncertainty analysis by examining the activations in the networks' final layer. Ideally, a confidence close to 1 suggests that the input sample is located in a region of the feature space familiar to the network, while a confidence near 0 indicates unfamiliarity. Consequently, we can disregard low-confidence predictions to minimize the risk of incorrect classifications. 
In Bayesian statistics, uncertainties are categorized as epistemic and aleatoric~\cite{NIPS2017_2650d608}. Epistemic uncertainty pertains to the model parameters, indicating that the model parameters have learned the under-represented region of the data. Aleatoric uncertainty, on the other hand, relates to the data itself, suggesting that the data is inherently noisy.

In this study, we employed the entropy of the neural output as a foundation for the confidence score, serving as a comprehensive measure of both epistemic and aleatoric uncertainty. Mathematically, given the network outputs $y_i, i \in [1, N]$ for classifying $N$ classes, the entropy is calculated as:
$$ H = \sum_{i=1}^N SoftMax(y_i) \times log_N(min(1, SoftMax(y_i) + \epsilon)),$$

Here, $min(\cdot)$ and $\epsilon$ are incorporated to circumvent numerical instability, and $SoftMax$ is conventionally defined as:
$$SoftMax(y_i) = \frac{e^{y_i}}{\sum_{j=1}^N e^{y_j}}$$

For $N=2$, $H$ aligns with the classical Shannon entropy computed using bits as information units. Generally, it always lies within $[0, 1]$, allowing the computation of a confidence score as $1 - H$.

We conducted all experiments for both the binary classification task, which involves distinguishing between ``musically relevant'' and ``musically irrelevant'' blobs, and the multi-class classification task, which involves differentiating among the 14 classes of objects.

\section{Results}
\label{sec:results}

\begin{table}
\caption{Neural network performances for the \textbf{binary} classification task. Note that the balanced accuracy is equivalent to the average recall. The best average values for each measure are highlighted in bold. The symbol ``-'' means that no data was retained for that class at that level of confidence.}
\label{tab:binary}
\resizebox{0.8\textwidth}{!}{% THIS IS TO AUTOMATICALLY RESIZE!
\begin{tabular}{|cccccccccccccccc|}
\hline
\multicolumn{16}{|c|}{\textbf{DenseNet}}                                                                                                                                                                                                                                                                                                                                                                                                            \\ \hline
\multicolumn{1}{|c|}{\textbf{}}                          & \multicolumn{5}{c|}{\textbf{Precision}}                                                                                                        & \multicolumn{5}{c|}{\textbf{Recall}}                                                                                                           & \multicolumn{5}{c|}{\textbf{F1-score}}                                                                                    \\
\multicolumn{1}{|c|}{\textit{\textbf{Confidence level}}} & \textit{\textbf{0\%}} & \textit{\textbf{25\%}} & \textit{\textbf{50\%}} & \textit{\textbf{75\%}} & \multicolumn{1}{c|}{\textit{\textbf{90\%}}} & \textit{\textbf{0\%}} & \textit{\textbf{25\%}} & \textit{\textbf{50\%}} & \textit{\textbf{75\%}} & \multicolumn{1}{c|}{\textit{\textbf{90\%}}} & \textit{\textbf{0\%}} & \textit{\textbf{25\%}} & \textit{\textbf{50\%}} & \textit{\textbf{75\%}} & \textit{\textbf{90\%}} \\ \hline
\multicolumn{1}{|c|}{Irrelevant}                         & 0.98                  & 0.99                   & 0.99                   & 1.00                   & \multicolumn{1}{c|}{1.00}                   & 0.82                  & 0.89                   & 0.93                   & 0.97                   & \multicolumn{1}{c|}{0.99}                   & 0.89                  & 0.94                   & 0.96                   & 0.98                   & 1.00                   \\
\multicolumn{1}{|c|}{Relevant}                           & 0.44                  & 0.56                   & 0.66                   & 0.77                   & \multicolumn{1}{c|}{0.87}                   & 0.87                  & 0.94                   & 0.95                   & 0.98                   & \multicolumn{1}{c|}{0.98}                   & 0.58                  & 0.70                   & 0.78                   & 0.86                   & 0.92                   \\ \hline
\multicolumn{1}{|c|}{\textit{Average}}                   & \textit{0.71}         & \textit{0.78}          & \textit{0.83}          & \textit{0.88}          & \multicolumn{1}{c|}{\textit{0.93}}          & \textit{0.85}         & \textit{0.91}          & \textit{0.94}          & \textit{0.97}          & \multicolumn{1}{c|}{\textit{\textbf{0.98}}} & \textit{0.74}         & \textit{0.82}          & \textit{0.87}          & \textit{0.92}          & \textit{\textbf{0.96}} \\ \hline
\multicolumn{16}{|c|}{\textbf{ResNet}}                                                                                                                                                                                                                                                                                                                                                                                                                                                 \\ \hline
\multicolumn{1}{|c|}{\textbf{}}                          & \multicolumn{5}{c|}{\textbf{Precision}}                                                                                                        & \multicolumn{5}{c|}{\textbf{Recall}}                                                                                                           & \multicolumn{5}{c|}{\textbf{F1-score}}                                                                                    \\
\multicolumn{1}{|c|}{\textit{\textbf{Confidence level}}} & \textit{\textbf{0\%}} & \textit{\textbf{25\%}} & \textit{\textbf{50\%}} & \textit{\textbf{75\%}} & \multicolumn{1}{c|}{\textit{\textbf{90\%}}} & \textit{\textbf{0\%}} & \textit{\textbf{25\%}} & \textit{\textbf{50\%}} & \textit{\textbf{75\%}} & \multicolumn{1}{c|}{\textit{\textbf{90\%}}} & \textit{\textbf{0\%}} & \textit{\textbf{25\%}} & \textit{\textbf{50\%}} & \textit{\textbf{75\%}} & \textit{\textbf{90\%}} \\ \hline
\multicolumn{1}{|c|}{Irrelevant}                         & 0.98                  & 0.99                   & 0.99                   & 1.00                   & \multicolumn{1}{c|}{1.00}                   & 0.82                  & 0.90                   & 0.95                   & 0.98                   & \multicolumn{1}{c|}{1.00}                   & 0.89                  & 0.94                   & 0.97                   & 0.99                   & 1.00                   \\
\multicolumn{1}{|c|}{Relevant}                           & 0.45                  & 0.58                   & 0.69                   & 0.82                   & \multicolumn{1}{c|}{0.90}                   & 0.87                  & 0.94                   & 0.94                   & 0.97                   & \multicolumn{1}{c|}{0.95}                   & 0.59                  & 0.72                   & 0.79                   & 0.89                   & 0.92                   \\ \hline
\multicolumn{1}{|c|}{\textit{Average}}                   & \textit{0.71}         & \textit{0.79}          & \textit{0.84}          & \textit{0.91}          & \multicolumn{1}{c|}{\textit{\textbf{0.95}}} & \textit{0.85}         & \textit{0.92}          & \textit{0.95}          & \textit{\textbf{0.98}} & \multicolumn{1}{c|}{\textit{0.97}}          & \textit{0.74}         & \textit{0.83}          & \textit{0.88}          & \textit{0.94}          & \textit{\textbf{0.96}} \\ \hline
\multicolumn{16}{|c|}{\textbf{GoogleNet}}                                                                                                                                                                                                                                                                                                                                                                                                                                              \\ \hline
\multicolumn{1}{|c|}{\textbf{}}                          & \multicolumn{5}{c|}{\textbf{Precision}}                                                                                                        & \multicolumn{5}{c|}{\textbf{Recall}}                                                                                                           & \multicolumn{5}{c|}{\textbf{F1-score}}                                                                                    \\
\multicolumn{1}{|c|}{\textit{\textbf{Confidence level}}} & \textit{\textbf{0\%}} & \textit{\textbf{25\%}} & \textit{\textbf{50\%}} & \textit{\textbf{75\%}} & \multicolumn{1}{c|}{\textit{\textbf{90\%}}} & \textit{\textbf{0\%}} & \textit{\textbf{25\%}} & \textit{\textbf{50\%}} & \textit{\textbf{75\%}} & \multicolumn{1}{c|}{\textit{\textbf{90\%}}} & \textit{\textbf{0\%}} & \textit{\textbf{25\%}} & \textit{\textbf{50\%}} & \textit{\textbf{75\%}} & \textit{\textbf{90\%}} \\ \hline
\multicolumn{1}{|c|}{Irrelevant}                         & 0.98                  & 0.99                   & 0.99                   & 1.00                   & \multicolumn{1}{c|}{1.00}                   & 0.82                  & 0.90                   & 0.94                   & 0.97                   & \multicolumn{1}{c|}{1.00}                   & 0.89                  & 0.94                   & 0.97                   & 0.99                   & 1.00                   \\
\multicolumn{1}{|c|}{Relevant}                           & 0.45                  & 0.58                   & 0.68                   & 0.78                   & \multicolumn{1}{c|}{0.90}                   & 0.88                  & 0.94                   & 0.95                   & 0.98                   & \multicolumn{1}{c|}{0.97}                   & 0.59                  & 0.71                   & 0.79                   & 0.87                   & 0.93                   \\ \hline
\multicolumn{1}{|c|}{\textit{Average}}                   & \textit{0.71}         & \textit{0.78}          & \textit{0.84}          & \textit{0.89}          & \multicolumn{1}{c|}{\textit{\textbf{0.95}}} & \textit{0.85}         & \textit{0.92}          & \textit{0.95}          & \textit{\textbf{0.98}} & \multicolumn{1}{c|}{\textit{\textbf{0.98}}} & \textit{0.74}         & \textit{0.83}          & \textit{0.88}          & \textit{0.93}          & \textit{\textbf{0.96}} \\ \hline
\end{tabular}
}
\end{table}

\begin{figure}
    \centering
    \includegraphics[width=0.7\textwidth]{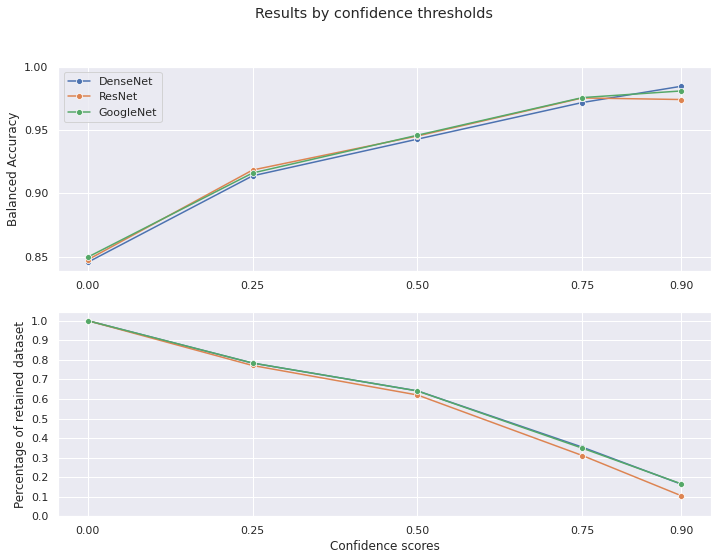}
    \Description{Trend of the balanced accuracy and percentage of retained data for various level of confidences for the \textbf{binary} task.}

    \caption{Trend of the balanced accuracy and percentage of retained test data for various level of confidences for the \textbf{binary} task.}
    \label{fig:binary}
\end{figure}

For the binary classification task, all three deep learning models achieved a balanced accuracy of 85\% and an f1-score of 74\%, as detailed in Table~\ref{tab:binary}. 
In contrast, the constant predictor yielded a balanced accuracy of 50\% and an f1-score of 46\%, underperforming the random guessing.

By considering varying confidence levels, we observed a consistent monotonic trend for both accuracy and the proportion of retained test data. This indicates that higher accuracies can be attained by predicting fewer samples, as illustrated in Fig.~\ref{fig:binary}. 
For example, with GoogleNet, a balanced accuracy of 95\% and an f1-score of 88\% can be achieved by retaining only samples with a confidence exceeding 50\%, which constitutes 64\% of the test set.

\begin{table}
\caption{Neural network performances for the \textbf{multiclass} task. Note that the balanced accuracy is equivalent to the average recall. The best average values for each measure are highlighted in bold. The symbol ``-'' means that no data was retained for that class at that level of confidence.}
\label{tab:multiclass}
\resizebox{\textwidth}{!}{% THIS IS TO AUTOMATICALLY RESIZE!
\begin{tabular}{|cccccccccccccccc|}
\hline
\multicolumn{16}{|c|}{\textbf{DenseNet}}                                                                                                                                                                                                                                                                                                                                                                                                                                                             \\ \hline
\multicolumn{1}{|c|}{\textbf{}}                                        & \multicolumn{5}{c|}{\textbf{Precision}}                                                                                                        & \multicolumn{5}{c|}{\textbf{Recall}}                                                                                                           & \multicolumn{5}{c|}{\textbf{F1-score}}                                                                                    \\
\multicolumn{1}{|c|}{\textit{Confidence level}}                        & \textit{\textbf{0\%}} & \textit{\textbf{25\%}} & \textit{\textbf{50\%}} & \textit{\textbf{75\%}} & \multicolumn{1}{c|}{\textit{\textbf{90\%}}} & \textit{\textbf{0\%}} & \textit{\textbf{25\%}} & \textit{\textbf{50\%}} & \textit{\textbf{75\%}} & \multicolumn{1}{c|}{\textit{\textbf{90\%}}} & \textit{\textbf{0\%}} & \textit{\textbf{25\%}} & \textit{\textbf{50\%}} & \textit{\textbf{75\%}} & \textit{\textbf{90\%}} \\ \hline
\multicolumn{1}{|c|}{Single note (with at least the head)}             & 0.11                  & 0.16                   & 0.00                   & 0.00                   & \multicolumn{1}{c|}{-}                      & 0.31                  & 0.27                   & 0.00                   & 0.00                   & \multicolumn{1}{c|}{-}                      & 0.16                  & 0.20                   & 0.00                   & 0.00                   & -                      \\
\multicolumn{1}{|c|}{Manuscript Text}                                  & 0.42                  & 0.57                   & 0.86                   & 1.00                   & \multicolumn{1}{c|}{0.00}                   & 0.37                  & 0.46                   & 0.46                   & 0.13                   & \multicolumn{1}{c|}{0.00}                   & 0.40                  & 0.51                   & 0.60                   & 0.24                   & 0.00                   \\
\multicolumn{1}{|c|}{Remaining}                                        & 0.15                  & 0.21                   & 0.00                   & 0.00                   & \multicolumn{1}{c|}{0.00}                   & 0.21                  & 0.18                   & 0.00                   & 0.00                   & \multicolumn{1}{c|}{0.00}                   & 0.17                  & 0.19                   & 0.00                   & 0.00                   & 0.00                   \\
\multicolumn{1}{|c|}{Printed Text}                                     & 0.56                  & 0.69                   & 0.85                   & 0.92                   & \multicolumn{1}{c|}{0.94}                   & 0.67                  & 0.86                   & 0.97                   & 0.99                   & \multicolumn{1}{c|}{1.00}                   & 0.61                  & 0.77                   & 0.90                   & 0.96                   & 0.97                   \\
\multicolumn{1}{|c|}{Key(s) (whole(s) or nearly)}                      & 0.50                  & 0.72                   & 0.91                   & 1.00                   & \multicolumn{1}{c|}{1.00}                   & 0.77                  & 0.94                   & 1.00                   & 1.00                   & \multicolumn{1}{c|}{1.00}                   & 0.61                  & 0.82                   & 0.95                   & 1.00                   & 1.00                   \\
\multicolumn{1}{|c|}{Blur or multiple categories (no music signs)}    & 0.26                  & 0.30                   & 0.59                   & 0.33                   & \multicolumn{1}{c|}{0.00}                   & 0.49                  & 0.58                   & 0.50                   & 0.05                   & \multicolumn{1}{c|}{0.00}                   & 0.34                  & 0.39                   & 0.54                   & 0.09                   & 0.00                   \\
\multicolumn{1}{|c|}{Multiple notes or chords}                         & 0.29                  & 0.38                   & 0.79                   & 0.00                   & \multicolumn{1}{c|}{0.00}                   & 0.38                  & 0.59                   & 0.56                   & 0.00                   & \multicolumn{1}{c|}{0.00}                   & 0.33                  & 0.46                   & 0.65                   & 0.00                   & 0.00                   \\
\multicolumn{1}{|c|}{Page border}                                      & 0.95                  & 0.95                   & 0.98                   & 0.99                   & \multicolumn{1}{c|}{0.99}                   & 0.77                  & 0.89                   & 0.98                   & 1.00                   & \multicolumn{1}{c|}{0.99}                   & 0.85                  & 0.92                   & 0.98                   & 0.99                   & 0.99                   \\
\multicolumn{1}{|c|}{Other (without musical markings)}                 & 0.23                  & 0.31                   & 0.43                   & 0.75                   & \multicolumn{1}{c|}{0.00}                   & 0.18                  & 0.27                   & 0.47                   & 0.50                   & \multicolumn{1}{c|}{0.00}                   & 0.20                  & 0.29                   & 0.45                   & 0.60                   & 0.00                   \\
\multicolumn{1}{|c|}{Erasure}                                          & 0.26                  & 0.34                   & 0.67                   & 0.00                   & \multicolumn{1}{c|}{0.00}                   & 0.41                  & 0.39                   & 0.27                   & 0.00                   & \multicolumn{1}{c|}{0.00}                   & 0.32                  & 0.37                   & 0.39                   & 0.00                   & 0.00                   \\
\multicolumn{1}{|c|}{More categories (with at least one musical sign)} & 0.28                  & 0.36                   & 0.00                   & 0.00                   & \multicolumn{1}{c|}{0.00}                   & 0.11                  & 0.09                   & 0.00                   & 0.00                   & \multicolumn{1}{c|}{0.00}                   & 0.16                  & 0.14                   & 0.00                   & 0.00                   & 0.00                   \\ \hline
\multicolumn{1}{|c|}{\textit{Average}}                                 & \textit{0.36}         & \textit{0.45}          & \textit{0.55}          & \textit{0.45}          & \multicolumn{1}{c|}{\textit{0.29}}          & \textit{0.42}         & \textit{0.50}          & \textit{0.47}          & \textit{0.33}          & \multicolumn{1}{c|}{\textit{0.30}}          & \textit{0.38}         & \textit{0.46}          & \textit{0.50}          & \textit{0.35}          & \textit{0.30}          \\ \hline
\multicolumn{16}{|c|}{\textbf{ResNet}}                                                                                                                                                                                                                                                                                                                                                                                                                                                               \\ \hline
\multicolumn{1}{|c|}{\textbf{}}                                        & \multicolumn{5}{c|}{\textbf{Precision}}                                                                                                        & \multicolumn{5}{c|}{\textbf{Recall}}                                                                                                           & \multicolumn{5}{c|}{\textbf{F1-score}}                                                                                    \\
\multicolumn{1}{|c|}{\textit{Confidence level}}                        & \textit{\textbf{0\%}} & \textit{\textbf{25\%}} & \textit{\textbf{50\%}} & \textit{\textbf{75\%}} & \multicolumn{1}{c|}{\textit{\textbf{90\%}}} & \textit{\textbf{0\%}} & \textit{\textbf{25\%}} & \textit{\textbf{50\%}} & \textit{\textbf{75\%}} & \multicolumn{1}{c|}{\textit{\textbf{90\%}}} & \textit{\textbf{0\%}} & \textit{\textbf{25\%}} & \textit{\textbf{50\%}} & \textit{\textbf{75\%}} & \textit{\textbf{90\%}} \\ \hline
\multicolumn{1}{|c|}{Single note (with at least the head)}             & 0.12                  & 0.16                   & 0.00                   & 0.00                   & \multicolumn{1}{c|}{-}                      & 0.33                  & 0.30                   & 0.00                   & 0.00                   & \multicolumn{1}{c|}{-}                      & 0.18                  & 0.21                   & 0.00                   & 0.00                   & -                      \\
\multicolumn{1}{|c|}{Manuscript Text}                                  & 0.44                  & 0.56                   & 0.76                   & 1.00                   & \multicolumn{1}{c|}{0.00}                   & 0.38                  & 0.50                   & 0.50                   & 0.13                   & \multicolumn{1}{c|}{0.00}                   & 0.41                  & 0.53                   & 0.60                   & 0.24                   & 0.00                   \\
\multicolumn{1}{|c|}{Remaining}                                        & 0.14                  & 0.19                   & 0.00                   & 0.00                   & \multicolumn{1}{c|}{0.00}                   & 0.15                  & 0.08                   & 0.00                   & 0.00                   & \multicolumn{1}{c|}{0.00}                   & 0.15                  & 0.11                   & 0.00                   & 0.00                   & 0.00                   \\
\multicolumn{1}{|c|}{Printed Text}                                     & 0.61                  & 0.70                   & 0.87                   & 0.93                   & \multicolumn{1}{c|}{0.95}                   & 0.65                  & 0.83                   & 0.97                   & 0.99                   & \multicolumn{1}{c|}{1.00}                   & 0.63                  & 0.76                   & 0.91                   & 0.96                   & 0.98                   \\
\multicolumn{1}{|c|}{Key(s) (whole(s) or nearly)}                      & 0.38                  & 0.58                   & 0.85                   & 0.95                   & \multicolumn{1}{c|}{0.99}                   & 0.78                  & 0.94                   & 1.00                   & 1.00                   & \multicolumn{1}{c|}{1.00}                   & 0.51                  & 0.72                   & 0.92                   & 0.97                   & 1.00                   \\
\multicolumn{1}{|c|}{Blur or multiple categories (no music signs)}    & 0.26                  & 0.32                   & 0.65                   & 1.00                   & \multicolumn{1}{c|}{0.00}                   & 0.42                  & 0.49                   & 0.39                   & 0.10                   & \multicolumn{1}{c|}{0.00}                   & 0.32                  & 0.38                   & 0.48                   & 0.18                   & 0.00                   \\
\multicolumn{1}{|c|}{Multiple notes or chords}                         & 0.28                  & 0.34                   & 0.87                   & 0.00                   & \multicolumn{1}{c|}{0.00}                   & 0.37                  & 0.53                   & 0.81                   & 0.00                   & \multicolumn{1}{c|}{0.00}                   & 0.32                  & 0.42                   & 0.84                   & 0.00                   & 0.00                   \\
\multicolumn{1}{|c|}{Page border}                                      & 0.95                  & 0.95                   & 0.98                   & 0.99                   & \multicolumn{1}{c|}{1.00}                   & 0.74                  & 0.88                   & 0.97                   & 0.99                   & \multicolumn{1}{c|}{0.98}                   & 0.83                  & 0.91                   & 0.98                   & 0.99                   & 0.99                   \\
\multicolumn{1}{|c|}{Other (without musical markings)}                 & 0.16                  & 0.20                   & 0.36                   & 0.71                   & \multicolumn{1}{c|}{0.00}                   & 0.22                  & 0.29                   & 0.53                   & 0.80                   & \multicolumn{1}{c|}{0.00}                   & 0.18                  & 0.24                   & 0.43                   & 0.75                   & 0.00                   \\
\multicolumn{1}{|c|}{Erasure}                                          & 0.24                  & 0.34                   & 0.56                   & 1.00                   & \multicolumn{1}{c|}{-}                      & 0.42                  & 0.43                   & 0.35                   & 0.38                   & \multicolumn{1}{c|}{-}                      & 0.31                  & 0.38                   & 0.43                   & 0.55                   & -                      \\
\multicolumn{1}{|c|}{More categories (with at least one musical sign)} & 0.32                  & 0.44                   & 0.00                   & 0.00                   & \multicolumn{1}{c|}{-}                      & 0.18                  & 0.19                   & 0.00                   & 0.00                   & \multicolumn{1}{c|}{-}                      & 0.23                  & 0.27                   & 0.00                   & 0.00                   & -                      \\ \hline
\multicolumn{1}{|c|}{\textit{Average}}                                 & \textit{0.35}         & \textit{0.43}          & \textit{0.54}          & \textit{\textbf{0.60}} & \multicolumn{1}{c|}{\textit{0.37}}          & \textit{0.42}         & \textit{\textbf{0.50}} & \textit{\textbf{0.50}} & \textit{0.40}          & \multicolumn{1}{c|}{\textit{0.37}}          & \textit{0.37}         & \textit{0.45}          & \textit{\textbf{0.51}} & \textit{0.42}          & \textit{0.37}          \\ \hline
\multicolumn{16}{|c|}{\textbf{GoogleNet}}                                                                                                                                                                                                                                                                                                                                                                                                                                                            \\ \hline
\multicolumn{1}{|c|}{\textbf{}}                                        & \multicolumn{5}{c|}{\textbf{Precision}}                                                                                                        & \multicolumn{5}{c|}{\textbf{Recall}}                                                                                                           & \multicolumn{5}{c|}{\textbf{F1-score}}                                                                                    \\
\multicolumn{1}{|c|}{\textit{Confidence level}}                        & \textit{\textbf{0\%}} & \textit{\textbf{25\%}} & \textit{\textbf{50\%}} & \textit{\textbf{75\%}} & \multicolumn{1}{c|}{\textit{\textbf{90\%}}} & \textit{\textbf{0\%}} & \textit{\textbf{25\%}} & \textit{\textbf{50\%}} & \textit{\textbf{75\%}} & \multicolumn{1}{c|}{\textit{\textbf{90\%}}} & \textit{\textbf{0\%}} & \textit{\textbf{25\%}} & \textit{\textbf{50\%}} & \textit{\textbf{75\%}} & \textit{\textbf{90\%}} \\ \hline
\multicolumn{1}{|c|}{Single note (with at least the head)}             & 0.11                  & 0.15                   & 0.00                   & -                      & \multicolumn{1}{c|}{-}                      & 0.34                  & 0.31                   & 0.00                   & -                      & \multicolumn{1}{c|}{-}                      & 0.17                  & \textbf{0.21}          & 0.00                   & -                      & -                      \\
\multicolumn{1}{|c|}{Manuscript Text}                                  & 0.43                  & 0.57                   & 1.00                   & 0.00                   & \multicolumn{1}{c|}{0.00}                   & 0.34                  & 0.41                   & 0.42                   & 0.00                   & \multicolumn{1}{c|}{0.00}                   & 0.38                  & 0.48                   & \textbf{0.59}          & 0.00                   & 0.00                   \\
\multicolumn{1}{|c|}{Remaining}                                        & 0.14                  & 0.18                   & 0.00                   & 0.00                   & \multicolumn{1}{c|}{0.00}                   & 0.15                  & 0.08                   & 0.00                   & 0.00                   & \multicolumn{1}{c|}{0.00}                   & 0.15                  & 0.11                   & 0.00                   & 0.00                   & 0.00                   \\
\multicolumn{1}{|c|}{Printed Text}                                     & 0.58                  & 0.69                   & 0.83                   & 0.92                   & \multicolumn{1}{c|}{0.95}                   & 0.67                  & 0.84                   & 0.96                   & 0.99                   & \multicolumn{1}{c|}{1.00}                   & 0.62                  & 0.76                   & 0.89                   & 0.95                   & 0.97                   \\
\multicolumn{1}{|c|}{Key(s) (whole(s) or nearly)}                      & 0.46                  & 0.70                   & 0.92                   & 1.00                   & \multicolumn{1}{c|}{1.00}                   & 0.79                  & 0.93                   & 0.99                   & 1.00                   & \multicolumn{1}{c|}{1.00}                   & 0.58                  & 0.80                   & 0.96                   & 1.00                   & 1.00                   \\
\multicolumn{1}{|c|}{Blur or multiple categories (no music signs)}    & 0.28                  & 0.32                   & 0.67                   & 1.00                   & \multicolumn{1}{c|}{0.00}                   & 0.42                  & 0.49                   & 0.36                   & 0.04                   & \multicolumn{1}{c|}{0.00}                   & 0.33                  & 0.39                   & 0.47                   & 0.07                   & 0.00                   \\
\multicolumn{1}{|c|}{Multiple notes or chords}                         & 0.27                  & 0.34                   & 0.71                   & 0.00                   & \multicolumn{1}{c|}{0.00}                   & 0.43                  & 0.64                   & 0.76                   & 0.00                   & \multicolumn{1}{c|}{0.00}                   & 0.33                  & 0.44                   & 0.73                   & 0.00                   & 0.00                   \\
\multicolumn{1}{|c|}{Page border}                                      & 0.94                  & 0.95                   & 0.98                   & 0.99                   & \multicolumn{1}{c|}{0.99}                   & 0.78                  & 0.89                   & 0.99                   & 1.00                   & \multicolumn{1}{c|}{1.00}                   & 0.85                  & 0.92                   & 0.98                   & 0.99                   & 0.99                   \\
\multicolumn{1}{|c|}{Other (without musical markings)}                 & 0.20                  & 0.24                   & 0.50                   & 0.80                   & \multicolumn{1}{c|}{1.00}                   & 0.18                  & 0.26                   & 0.56                   & 0.57                   & \multicolumn{1}{c|}{1.00}                   & 0.19                  & 0.25                   & 0.53                   & 0.67                   & 1.00                   \\
\multicolumn{1}{|c|}{Erasure}                                          & 0.26                  & 0.36                   & 0.61                   & 1.00                   & \multicolumn{1}{c|}{0.00}                   & 0.44                  & 0.48                   & 0.34                   & 0.20                   & \multicolumn{1}{c|}{0.00}                   & 0.33                  & 0.42                   & 0.44                   & 0.33                   & 0.00                   \\
\multicolumn{1}{|c|}{More categories (with at least one musical sign)} & 0.28                  & 0.37                   & 0.00                   & 0.00                   & \multicolumn{1}{c|}{0.00}                   & 0.15                  & 0.11                   & 0.00                   & 0.00                   & \multicolumn{1}{c|}{0.00}                   & 0.20                  & 0.17                   & 0.00                   & 0.00                   & 0.00                   \\ \hline
\multicolumn{1}{|c|}{\textit{Average}}                                 & \textit{0.36}         & \textit{0.44}          & \textit{0.56}          & \textit{0.57}          & \multicolumn{1}{c|}{\textit{0.39}}          & \textit{0.43}         & \textit{\textbf{0.50}} & \textit{0.49}          & \textit{0.38}          & \multicolumn{1}{c|}{\textit{0.40}}          & \textit{0.38}         & \textit{0.45}          & \textit{\textbf{0.51}} & \textit{0.40}          & \textit{0.40}          \\ \hline
\end{tabular}
}

\end{table}

\begin{figure}
    \centering
    \includegraphics[width=0.7\textwidth]{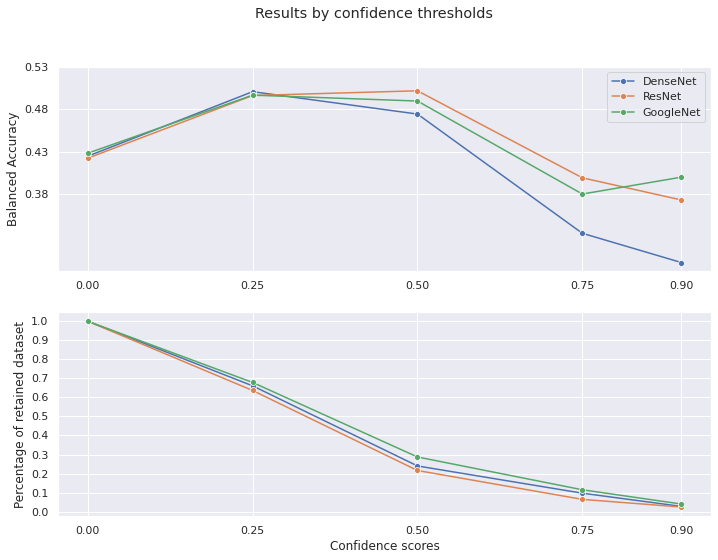}
    \Description{Trend of the balanced accuracy and percentage of retained data for various level of confidences for the \textbf{multiclass} classification task.}
    \caption{Trend of the balanced accuracy and percentage of retained data for various level of confidences for the \textbf{multiclass} classification task.}
    \label{fig:multiclass}
\end{figure}

In the multi-class classification task, the deep learning models achieved a balanced accuracy of 43\% and an f1-score of 38\%. For comparison, random guessing and constant predictors were used as baselines, yielding a balanced accuracy of 9\% and an f1-score of 6\%.

Upon considering confidence levels, we observed a convex accuracy curve with a peak between 25\% and 50\% for the three models. Specifically, GoogleNet and ResNet can achieve a balanced accuracy and f1-score of 50\% and 51\% respectively, by retaining 68\% of the data, as depicted in Fig.~\ref{fig:multiclass}. For confidence levels larger than 0.5, only printed text, keys, and page border can be identified satisfactorily, with f1-scores near to 1. However, a little number of samples are misclassified, leading the a fall in balanced accuracy, which is computed as the arithmetic mean of the per-class recall.

The per-class f1-score values, presented in Table~\ref{tab:multiclass}, reveal that the models are generally more proficient at predicting the most common classes. Specifically, the classes ``Page Border'', ``Printed Text'', and ``Keys'' achieved an f1-score of 85\%, 63\%, and 61\% respectively. These results are particularly beneficial for reducing the number of samples requiring manual annotation for dataset expansion.

The AutoML classifier, while not reaching the accuracy of the neural networks, achieved a balanced accuracy of 37\% and an f1-score of 31\% in the multi-class case. In the binary classification, it achieved a maximum of 84\% and 70\% respectively. The resulting models, composed of large ensembles of random forests, gradient boosting, support vector machines, linear and quadratic discriminant analysis, coupled with various pre-processing steps, are described in detail in the notebooks available in the source code repository. These architectures generate large models of several gigabytes. However, existing tools cannot leverage GPU processing like neural network frameworks, making the training and inference of such AutoML models more memory and time-intensive than the neural transfer learning approach.

For all the aforementioned metrics, detailed values and per-class statistics can be found in the notebook in the source code repository.

\section{Conclusions}
\label{sec:conclusions}

This study presents a comprehensive methodology for OMR applied to historical and handwritten music scores, with a particular focus on the Ricordi Archive. This prestigious archive, housing significant musical manuscripts from eminent opera composers, has been digitized and meticulously annotated to generate a novel dataset of musical symbols. This dataset, along with the models and source code employed in our experiments, is publicly available, thereby contributing to the wider research community.\footnote{\url{https://zenodo.org/doi/10.5281/zenodo.11186095}}

We have addressed several OMR challenges by training and evaluating multiple neural classifiers to differentiate between these symbols. Three renowned deep learning classifiers, namely ResNet, DenseNet, and GoogleNet, were fine-tuned, and a robust AutoML approach was utilized as a baseline. The deep learning models demonstrated promising results, achieving a balanced accuracy of 85\% in the binary classification task. By leveraging the confidence of the models, even higher accuracies were attained.

The primary contribution of this work lies in the creation of a unique dataset of musical symbols derived from real-world annotated manuscripts, which can be utilized to train and evaluate OMR models. Additionally, our work outlines a comprehensive methodology for preprocessing, annotating, and classifying musical symbols, which can be replicated and expanded upon in future research.

Future work will involve using the trained models to annotate additional data, discarding irrelevant sub-images and focusing on images where the model exhibits low confidence. This strategy will enable automatic pixel-wise classification of all pages, followed by a focus on image regions with lower confidence. The ability to identify musical objects will facilitate the provision of more specific labels for such objects. This approach will significantly simplify the annotation of the full corpus, providing the research community with an updated version of the dataset.

\begin{acks}
We would like to express our sincere gratitude to the Ricordi Archive for granting permission to utilize the graphical materials included in this paper. In addition, we extend our appreciation for the fruitful mutual collaboration that has taken place over the last 20 years.
We gratefully acknowledge the support of NVIDIA Corp. with the donation of two Titan V GPUs.
\end{acks}

\bibliographystyle{ACM-Reference-Format}
\bibliography{bibliography,other}

\end{document}